\documentclass[12pt,a4paper,twoside]{article}

\usepackage{epsfig,cite}      
\usepackage{amsmath,amsfonts}

\newcommand{\C}{\mathbb C}

\newcommand{\R}{\mathbb R}  
\newcommand{\Z}{\mathbb Z}

\newcommand{\al}{\alpha}
\newcommand{\be}{\beta}
\newcommand{\ga}{\gamma}

\newcommand{\si}{\sigma}

\begin{document}
\title{Three-Dimensional Face Orientation and Gaze Detection from a Single Image}
\author{Jeremy Yirmeyahu Kaminski$^{1}$, Mina Teicher$^1$, \\
Dotan Knaan$^2$ and Adi Shavit$^2$.\\
$^1$ Department of Mathematics, \\
Bar-Ilan University, \\
Ramat-Gan, Israel. \\
$^2$ Gentech Corporation, \\
Israel.
}
\date{}

\maketitle

\begin{abstract}
Gaze detection and head orientation are an important part of many advanced human-machine interaction applications. Many systems have been proposed for gaze detection. Typically, they require some form of user cooperation and calibration. Additionally, they may require multiple cameras and/or restricted head positions. We present a new approach for inference of both face orientation and gaze direction from a single image with no restrictions on the head position. Our algorithm is based on a face and eye model, deduced from anthropometric data. This approach allows us to use a single camera and requires no cooperation from the user. Using a single image avoids the complexities associated with of a multi-camera system. Evaluation tests show that our system is accurate, fast and can be used in a variety of applications, including ones where the user is unaware of the system.

\end{abstract}

\section{Introduction}

Eyes are a major way to acquire information about humans. Human attention, intention and even desire are closely related to gaze. As such, many applications require gaze detection. Instances are driver attention monitoring and human-computer interface for multimedia or medical purposes. The large number of proposed algorithms for this task proves that no solution is completely satisfying. Since quoting all the works related to the subject is impossible, we focus on some recent and important contributions. In \cite{Glenstrup}, one can find a good survey of some gaze detection techniques. In \cite{Ji-Yang-02,Perez-all-03}, one can find a stereo system for for gaze and face pose computation, which is particularly suitable for monitoring driver vigilance. Both systems are based on the two cameras, one being a narrow field camera (which provides a high resolution image of the eyes by tracking a small area) and the second being a large field camera (which tracks the whole face). Besides the computationally complex difficulties arising from multiple cameras and controlling these pan-tilt cameras, the system hardware quite costly. In \cite{Ohno-all-02}, a monocular system is presented, which uses a personal calibration process for each user and does not allow large head motions. Limiting the head motion is typical for systems that utilize only a single camera. \cite{Ohno-all-02} uses a (motorized) auto-focus lens to estimate the distance of the face from the camera. In \cite{Wang-all-03}, the eye gaze is computed by using the fact, that the iris contour, while being a circle in 3D is perspectively an ellipse in the image. The drawback in this approach is that a high resolution image of the iris area is necessary, which severely limits  the possible motions of the user, unless an additional wide-angle camera is used. 

In this paper we introduce a new approach with several advantages. The system is monocular, hence the difficulties associated with multiple cameras are avoided. The camera parameters are maintained constant in time. The system requires no personal calibration and the head is allowed to move freely.  This is achieved by using a model of the face, deduced from anthropometric features.

This approach, of a mechanically simple, automatic and non-intrusive system, allows eye-gazing to be used in a variety of applications where eye-gaze detection was not an option before. For example, such a system may be installed in mass produced cars. With the growing concern of car accidents, customers and regulators are demanding safer cars. Active sensors that may prevent accidents are actively perused. A non-intrusive, cheaply produced, one-size-fits-all eye-gazing system could monitor driver vigilance at all times. Drowsiness and inattention can immediately generate alarms. In conjunction with other active sensors, e.g. radar, obstacle detection, etc. the driver can be warned of an unnoticed hazard outside the car.

Psychophysical and psychological tests and experiments with uncooperative subjects such as children and/or primates, may also benefit from such a static (no moving parts) system, which allows the subject to focus solely on the task at hand while remaining oblivious to the eye-gaze system.

In conjunction with additional higher-level systems, a covert eye-gazing system may be useful in security applications. For example, monitoring the eye-gaze of ATM clients. In automated airport check–in counters, such a system may alert of suspiciously behaving individuals.

The paper is organized as follows. In section~\ref{sec::model}, we present the core of the paper, the face model that we use and how this model leads to the computation of the Euclidean face 3D orientation and position. Simulations are presented, that show the results are robust to error in both the model and the measurements. Section~\ref{sec::system} gives an overview of the system, and some experiments are presented.

\section{Face Model and Geometric Analysis}
\label{sec::model}

\subsection{Face Model}

\begin{figure}
\begin{center}
\psfig{figure=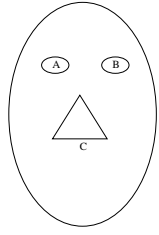,height={3.0cm}}
\caption{\protect\small The face model is essentially based on the fact that the triangle Eye-Nose Bottom-Eye is isosceles.}
\label{faceModel}
\end{center}
\end{figure}

Following the statistical data taken from \cite{Farkas-94}, we assume the following model for a generic human face. Let ${\bf A}$ and ${\bf B}$ be the centers of the eyes, and let ${\bf C}$ be the middle point between the nostrils. Then we assume the following model:
\begin{eqnarray}
\label{eqn::model1}
d({\bf A}, {\bf C}) = d({\bf B}, {\bf C}) \\
\label{eqn::model2}
d({\bf A}, {\bf B}) = r d({\bf A}, {\bf C}) \\
\label{eqn::model3}
d({\bf A}, {\bf B}) = 6.5 cm 
\end{eqnarray}
where $r=1.0833$. The two first equations allow computing the orientation of the face, while the third equation is necessary for computing the distance between the camera and the face. The face model is illustrated in figure~\ref{faceModel}.

\subsection{3D Face Orientation}

Let ${\bf M}$ be the camera matrix. All the computations are done in the coordinate system of the camera. Therefore the camera matrix has the following expression:
$$
\begin{array}{ccc}
{\bf M} & = & {\bf K} [{\bf I};{\bf 0}],
\end{array}
$$
where ${\bf K}$ is the matrix of internal parameters \cite{Hartley-Zisserman-00,Faugeras-Luong-01}.

Let $({\bf a},{\bf b}, {\bf c})$ be the projection of $({\bf A}, {\bf B} ,{\bf C})$ onto the image. In the equations below, the image points ${\bf a}, {\bf b}, {\bf c}$ are given by their projective coordinates in the image plane, while the 3D points ${\bf A}, {\bf B}, {\bf C}$ are given by their Euclidean coordinates in $\R^3$.  Given these notations, the projection equations are:
\begin{eqnarray}
{\bf a} & \sim & {\bf K A} \\
{\bf b} & \sim & {\bf K B} \\
{\bf c} & \sim & {\bf K C} 
\end{eqnarray} 
where $\sim$ means equality up to a scale factor. Therefore the 3D points are given by the following expressions:
\begin{eqnarray}
\label{eqn::eucl_A}
{\bf A } & = & \al {\bf K}^{-1} {\bf a} \\
\label{eqn::eucl_B}
{\bf B} & = & \be {\bf K}^{-1} {\bf b} \\
\label{eqn::eucl_C}
{\bf C} & = & \ga {\bf K}^{-1} {\bf c}
\end{eqnarray}
where $\al,\be,\ga$ are unknown scale factor. These could also be deduced by considering the points at infinity of the optical rays generated by the image points ${\bf a},{\bf b}, {\bf c}$ and the camera center. These points at infinity are simply given in projective coordinates by: $[{\bf K}^{-1}{\bf a},0]^t, [{\bf K}^{-1}{\bf b},0]^t, [{\bf K}^{-1}{\bf c},0]^t$. Then the points ${\bf A}, {\bf B}, {\bf C}$ are given in projective coordinates by $[\al {\bf K}^{-1}{\bf a},1]^t, [\be {\bf K}^{-1}{\bf b},1]^t, [\ga {\bf K}^{-1}{\bf c},1]^t$. These expressions naturally yields the equations~(\ref{eqn::eucl_A}),~(\ref{eqn::eucl_B}),~(\ref{eqn::eucl_C}) giving the Euclidean coordinates of the points.

Plugging these expressions of ${\bf A}, {\bf B}$ and ${\bf C}$ into the two first equations of the model~(\ref{eqn::model1}) and~(\ref{eqn::model2}), leads to two homogeneous quadratic equations in $\al,\be,\ga$:
\begin{eqnarray}
f(\al,\be,\ga) & = & 0 \\
g(\al,\be,\ga) & = & 0 
\end{eqnarray}

Thus finding the points ${\bf A}, {\bf B}$ and ${\bf C}$ is now reduced in finding the intersection of two conics in the projective plane. Moreover since no solution is on the line defined by $\ga = 0$ (since the nose of the user is not located at the camera center!), one can reduce the computation of the affine piece defined by $\ga =1$. Hence we shall now focus our attention on the following system:
\begin{eqnarray}
\label{eqn::aff1}
f(\al,\be,1) & = & 0 \\
\label{eqn::aff2}
g(\al,\be,1) & = & 0 
\end{eqnarray}

This system defines the intersection of two conics in the affine plane. The following subsection is devoted to the computation of the solutions of this system.

\subsection{Computing the Intersection of Conics in the Affine Plane}

For sake of completeness, we shall recall shortly one way of computing the solutions of the system above. For more details, see \cite{Sturmfels-02}. Consider first two
polynomials $f,g \in \C[x]$. The
resultant gives a way to know if the two polynomials have a common
root. Write the polynomials as follows:
$$
\left\{\begin{array}{ccc}
f & = & a_n x^n + ... + a_1 x + a_0 \\
g & = & b_p x^p + ... + b_1 x + b_0
\end{array}\right.
$$
The resultant of $f$ and $g$ is a polynomial $r$, which is a
combination of monomials in $\{a_i\}_{i=1,...,n}$ and
$\{b_j\}_{j=1,...,p}$ with coefficients in $\Z$, that is $r \in
\Z[a_i,b_j]$. The resultant $r$ vanish if and only if either $a_n$ or
$b_p$ is zero or the polynomials have a common root in $\C$. The
resultant can be computed as the determinant of a polynomial
matrix. There exist several matrices whose determinant is equal to the
resultant. The best known and simplest matrix is the so-called
Sylvester matrix, defined as follows:
$$
S(f,g) = \left[\begin{array}{cccccccccc}
        a_n     & 0      & 0      & ... & 0 & b_p     & 0   & 0 & ... & 0 \\
        a_{n-1} & a_n    & 0      & ... & 0 & b_{p-1} & b_p & 0 & ... & 0 \\
        \vdots  & \vdots & .      &     &   &         &     &   &&
                \end{array}\right]
$$
Therefore, we have:
$$
r(x) = det(Syl(f,g)).
$$
In addition to this expression which gives a practical way to compute
the resultant, there exists another formula of theoretical interest:
$$
r(x) = a_n b_p \Pi_{\al, \be} (x^f_\al - x^g_\be),
$$
where $x^f_\al$ are the roots of $f$ and $x^g_\be$ are those of $g$. It
can be shown that the resultant is a polynomial of degree $np$.

An important point is that the resultant is also defined and has the
same properties if the coefficients of the polynomials are not only
numbers but also polynomials in another variable. Hence, consider now
that $f,g \in \C[x,y]$ and write:
\begin{equation}
\label{equ1}
\left\{\begin{array}{ccc}
f & = & a_n(x) y^n + ... + a_1(x) y + a_0(x) \\
g & = & b_p(x) y^p + ... + b_1(x) y + b_0(x)
\end{array}\right.
\end{equation}
The question is now the following: given a value $x_0$ of $x$, do the
two polynomials $f(x_0,y)$ and $g(x_0,y)$ have a common root? The answer to this question is based on the computation of the resultant of $f$ and $g$ with respect to
$y$ (i.e. using the presentation given by(~\ref{equ1})) .
This is a univariate polynomial in $x$, denoted by $r(x) = res(f,g,y)$.

The resultant can be used in many contexts. For our purpose, we will
use it to compute the intersection points of two planar algebraic
curves. Consider the curve ${\cal C}_1$ (respectively ${\cal C}_2$)
defined as the set of points $(x,y)$
which are roots of $f(x,y)$ (respectively $g(x,y)$). We want to compute the
intersection of ${\cal C}_1$ and ${\cal C}_2$. Algebraically, this is
equivalent to compute the common roots of $f$ and $g$. Therefore, we
use the following procedure:
\begin{itemize}
\item Compute the resultant $r(x) = res(f,g,y) \in \C[x]$.
\item Find the roots of $r(x)$: $x_1,...,x_t$
\item For each $i=1,...,t$, compute the common roots of $f(x_i,y)$
and $g(x_i,y)$ in $\C[y]$: $y_{i1},...,y_{ik_i}$.
\item The intersection of ${\cal C}_1$ and ${\cal C}_2$ is therefore: \\
$(x_1,y_{11}),...,(x_1,y_{1k_1}),...,(x_t,y_{t1}),...,(x_t,y_{tk_t})$.
\end{itemize}

In our context, the resultant $r$ is polynomial of degree $4$ and so $t \leq 4$ and $k_i \leq 2$. To complete the picture, we just need to mention an efficient and
reliable way to compute the roots of a univariate polynomial. The
algorithm that we will describe is very efficient and robust for low
degree polynomials. Given a univariate polynomial $p(x) = a_n x^n +
... + a_1 x +a_0$, one can form the following matrix, called the {\it
companion matrix of $p$}:
$$
C(p) = \left[\begin{array}{ccccc}
        0       & 1     & 0     & ...   & 0 \\
        0       & 0     & 1     & ...   & 0 \\
        \vdots  & \vdots&       & \ddots&   \\
        -a_0/a_n& -a_1/a_n & -a_2/a_n & \ldots & -a_{n-1}/a_n
        \end{array}\right]
$$

A short computation shows that the characteristic polynomial of
$C(p)$ is equal to $-\frac{1}{a_n}p$. Thus the roots of $p$ are
exactly the eigenvalues of $C(p)$. This provides one practical way to
compute the roots of a univariate polynomial.

\subsection{3D Face Orientation}

Therefore, we solve the system $S$ defined by equations~(\ref{eqn::aff1}) and~(\ref{eqn::aff2}) using the approach presented above. By Bezout's theorem (or simply by looking at the degree of the resultant), we know that there are at most $4$ complex solutions to this system. Experiments show that system generated by the image of a human face has only two real roots. The ambiguity between these two roots is easily handled, since one solution leads to non realistic inter eyes distance. Let $(\al_0,\be_0)$ be the right solution. Then the points ${\bf A}, {\bf B}$ and ${\bf C}$ are known up to a unique scale factor. We shall denote ${\bf A}_0, {\bf B}_0$ and ${\bf C}_0$ the points obtained by the solution $(\al_0,\be_0)$, Thus we have the following expression:
\begin{eqnarray}
{\bf A }_0 & = & \al_0 {\bf K}^{-1} {\bf a} \\
{\bf B}_0 & = & \be_0 {\bf K}^{-1} {\bf b} \\
{\bf C}_0 & = &  {\bf K}^{-1} {\bf c}
\end{eqnarray}
Thus we have the following relations too: ${\bf A} = \ga{\bf A}_0$, ${\bf B} = \ga {\bf B}_0$ and ${\bf C} = \ga {\bf C}_0$.  

The computation of $\ga$ is done using the third model equation~(\ref{eqn::model3}). Once the face points are computed, one can compute the distance between the user's face and the camera and so the 3D orientation of the face. Indeed the normal to the plane defined by ${\bf A}, {\bf B}$ and ${\bf C}$ is given by:
$$
\overrightarrow{N} = \overrightarrow{AB} \wedge \overrightarrow{AC},
$$
where $\wedge$ is the cross product.

\subsection{Robustness to Errors in Model and Detection}
\label{subsec::robust}

In order to estimate the sensitivity of this algorithm to errors in model and in detection, we performed several simulations. As we shall detail in subsection~\ref{subsec::arch}, we use a rather high resolution camera. Therefore in the simulation, we start from the following setting:
\begin{itemize}
\item The focal length $f = 4000$ in pixels,
\item The principal point is at the image center,
\item The distance between the camera and the face is $60 cm$.
\end{itemize}

The simulations are done according to the following protocol. An artifical face, defined by three points in space, say ${\bf A}, {\bf B}$ and ${\bf C}$, is projected onto a known camera. Given a parameter $p$, we perform a perturbation of $p$ by a white gaussian noise of standard deviation $\si$. For each value of $\si$, we perform 100 random perturbations. For each value of $p$, that we obtain by this process, we compute the error in the 3D reconstruction, as the mean of the square errors.  

\begin{figure}
\begin{center}
\psfig{figure=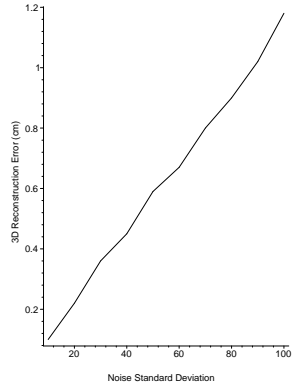,height={5cm}}
\caption{\protect\small Influence of the error in focal length.}
\label{focal-length}
\end{center}
\end{figure}

The first simulation (see figure~\ref{focal-length}) shows that the system is very robust to error in the estimation of the focal length, since for a noise with standard deviation of $100$ (in pixels), the reconstruction error is $1.2 cm$, meaning less than $1\%$ of the distance between the camera and the user.

\begin{figure}
\begin{center}
\psfig{figure=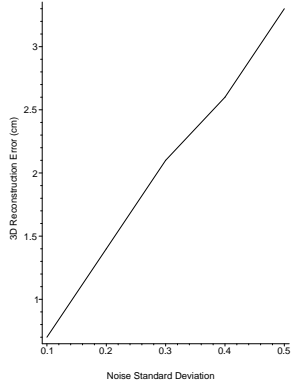,height={5cm}}
\caption{\protect\small Influence of the error in inter-eyes distance.}
\label{inter-eyes-dist}
\end{center}
\end{figure}

\begin{figure}
\begin{center}
\psfig{figure=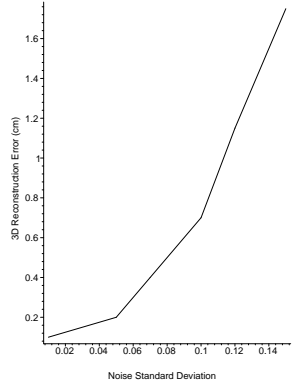,height={5cm}}
\caption{\protect\small Influence of the error in human ratio.}
\label{human-ratio}
\end{center}
\end{figure}

The next two simulation aim at measuring the influence of errors in model. First, the assumed inter-eyes distance is corrupted by a Gaussian white noise (figure~\ref{inter-eyes-dist}). The mean value is $6.5cm$ as mentionned in section~\ref{sec::model}. For a standard deviation of $0.5$, which represents an extreme anomaly with respect to the standard human morphology, the reconstruction error is about $3.3cm$, less than $2\%$ of the distance between the camera and the user. The influence of the human ratio $r$, as defined in equation~(\ref{eqn::model2}), is also tested, by adding a Gaussian white noise, centered at the "universal" value $1.0833$ (figure~\ref{human-ratio}). For a standard deviation $0.15$, which represents also a very strong anomaly, the reconstruction is $1.75cm$, just more than $1\%$ of the distance between the camera and the user. 

\begin{figure}
\begin{center}
\psfig{figure=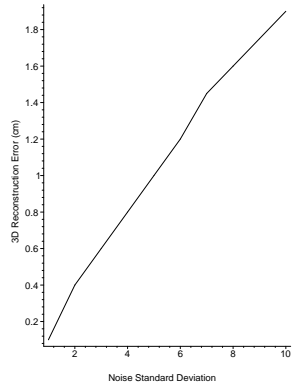,height={5cm}}
\caption{\protect\small Influence of the error in image points.}
\label{2dpoints}
\end{center}
\end{figure}

After measuring the influence of errors in camera calibration and model, the next step is evaluate the sensitivity to input data perturbation. The image points are corrupted by a Gaussian white noise (figure~\ref{2dpoints}). For noise of 10 pixels, which is a large error in detection, the reconstruction error is less $2cm$, about $1.15\%$ of the distance between the camera and the user.  

The accuracy of the system is mainly due to the fact that the focal length is high ($f=4000$ in pixels). Indeed when computing the optical rays generated by the image points, as in equations~(\ref{eqn::eucl_A},\ref{eqn::eucl_B},\ref{eqn::eucl_C}), we use the inverse of $K$, which is roughly equivalent to multiplying the image points coordinates by $1/f$. Hence the larger $f$ is, the less a detection error has an impact on the computation.

\section{Overview Of The System}
\label{sec::system}

\subsection{System Architecture}
\label{subsec::arch}

The main goal of this work was to create a non-intrusive gaze detection system, that would require no user cooperation while keeping the system complexity low. We use a high-resolution 15 fps, 1392x1040 video camera with a 25mm fixed-focus lens. This setup allows both a wide field of view, for a broad range of head positions, and high resolution images of the eyes. Since we can estimate the 3D head position from a single image, we can use a fixed focus lens instead of a motorized auto-focus lens. This makes the camera calibration simpler and the calibration of the internal parameters is done only once. The system uses an IR LED at a known position to illuminate the user's face.

\subsection{System Overview}

\begin{figure}
\begin{center}
\psfig{figure=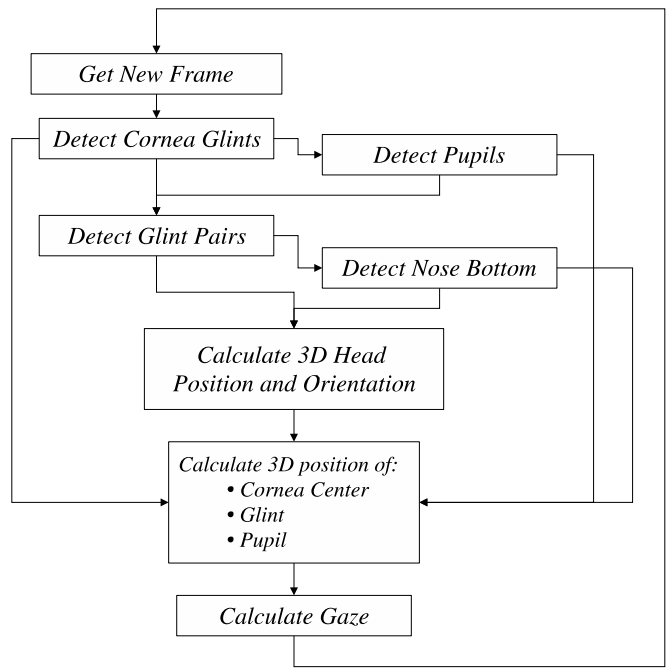,height={9.0cm}}
\caption{\protect\small The system flow chart, showing the different stages of the process.}
\label{flowChart}
\end{center}
\end{figure}

The general flow of the system is depicted in Figure~\ref{flowChart}. For every new frame the {\it glints}, the reflection of the LED light from the eye corneas as seen by the camera, are detected and their corresponding pupil is found. The search area for the nose is then defined, and the nose bottom is found. Given the two glints and nose position, we can reconstruct the complete Euclidean 3D face position and orientation relative to the camera, using the geometric algorithm presented in section~\ref{sec::model}. This reconstruction gives us the exact 3D position of the glints and pupils. Then, for each eye, the 3D cornea center is computed using the knowledge of LED position, as shown in figure~\ref{cornea-center}. This model is similar to the eye model used in~\cite{Ohno-all-02}.
The following sub-sections~\ref{subsec::features} and~\ref{subsec::gaze} will describe these stages in more detail.

\subsection{Feature Detection}
\label{subsec::features}

\subsubsection{Glint and Pupil Detection}

The detection of the glints is done in several steps. Glints appear as very bright dots in the image, usually at the highest possible grayscale values. Using a thresholding operation on the image yields multiple candidates for possible glints. Examples of other sources of similar characteristics are background lights, facial hair, teeth and eye-glasses lens and frames. We perform multiple filtering stages to identify the true glints. We filter these candidates by size, i.e. we select only the small dot-like ones. Next, we pair-up the remaining candidates and select only those glint-pairs that obey certain distance and angle rules and ranges. 

We next proceed to the detection of the pupils. The pupils serve two purposes. They are used to filter out incorrect glint pairs, and they are required for the calculation of the gaze direction in the later stages of the algorithm. Pupils appear as round or oval dark regions inside the eye and are very close to (or behind) the glints. We search for these dark regions around each of our detected glints. Glint pairs containing a glint around which no pupil was found are removed. This final glint filtering will usually leave us with the final true glint pair. Otherwise, we choose the top-most pair, as empirically, it was shown to be the correct one. 

\subsubsection{Nose Detection}

\begin{figure}
\begin{center}
\psfig{figure=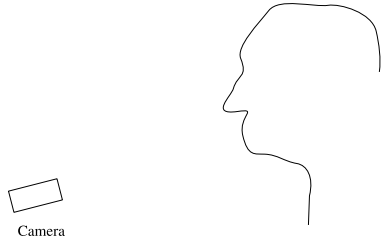,height={5.0cm}}
\caption{\protect\small The camera is viewing the eyes and the nostrils.}
\label{fig::face-camera}
\end{center}
\end{figure}

The detection of the nose-bottom, is done by searching for dark-bright-dark patterns in the area just below the eyes. Indeed, the nostrils appear as dark blobs in the image thanks to the relative position of the camera and the face as shown in figure~\ref{fig::face-camera}. The size and orientation of this search area is determined by the distance and orientation of the chosen glint-pair. Once dark-bright-dark patterns are found, we use connected component blob analysis on this region to identify only those dark blobs that obey certain size, shape, distance and relative angle rules that yield plausible nostrils. The nose bottom is selected as the point just between the two nostrils. 

\subsection{Gaze Detection}
\label{subsec::gaze}

\begin{figure}
\begin{center}
\psfig{figure=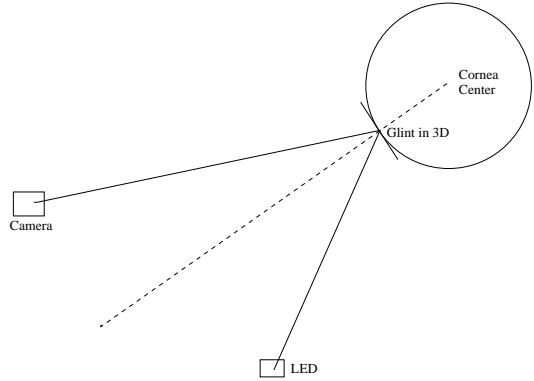,height={5.0cm}}
\caption{\protect\small The cornea center lies on the bisector of the angle defined by the LED, the glint point in 3D and the camera. Its exact location is given by the cornea radius, which is $77 mm$.}
\label{cornea-center}
\end{center}
\end{figure}

Given the glints and the bottom point of the nose are detected in the image, one can apply the geometric algorithm presented in section~\ref{sec::model} to compute the 3D face orientation. As seen in subsection~\ref{subsec::robust}, even if the glints are not exactly located in the center of the eye, the system returns an accurate answer. Then for each eye, the cornea center is computed using the knowledge of LED position, as shown in figure~\ref{cornea-center}. This model is similar to the one presented in \cite{Ohno-all-02}.

The gaze line is defined as being the line joining the cornea center and the pupil center in 3D.
The pupil center is first detected in the image and computed in 3D as follows. The distance between the pupil center and the cornea center is a known human anatomy data. It is equal to $0.45$ cm. Consider then a sphere $S$ centered at the cornea center, with radius equal to $0.45$ cm. The pupil center lies on the optical ray generated by its projection onto the image and the camera center. This ray intersects the sphere $S$ in two points. The closest of these points to the camera is the pupil center.  


\section{Experiments}

\begin{figure}
\psfig{figure=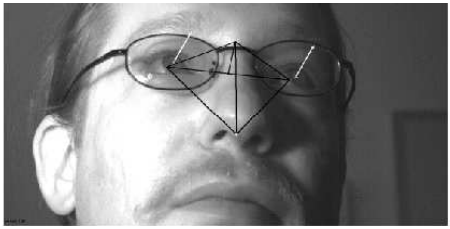,height={8cm}}
\caption{\protect\small The detected triangle, eyes' centers and the nose bottom, together with the gaze line.}
\label{fig::adi1}
\end{figure}

\begin{figure}
\psfig{figure=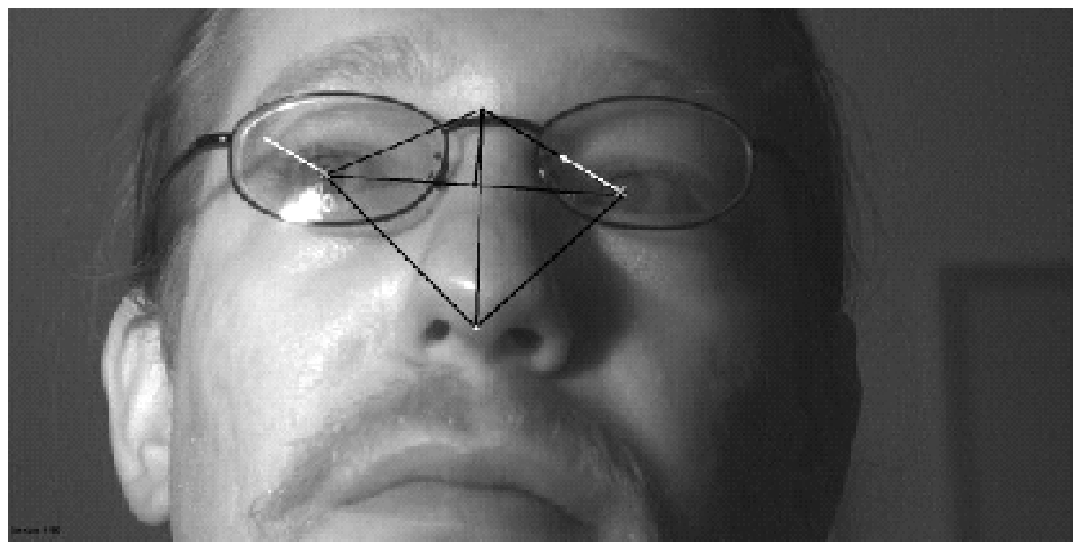,height={8cm}}
\caption{\protect\small The detected triangle, eyes' centers and the nose bottom, together with the gaze line.}
\label{fig::adi2}
\end{figure}

\begin{figure}
\psfig{figure=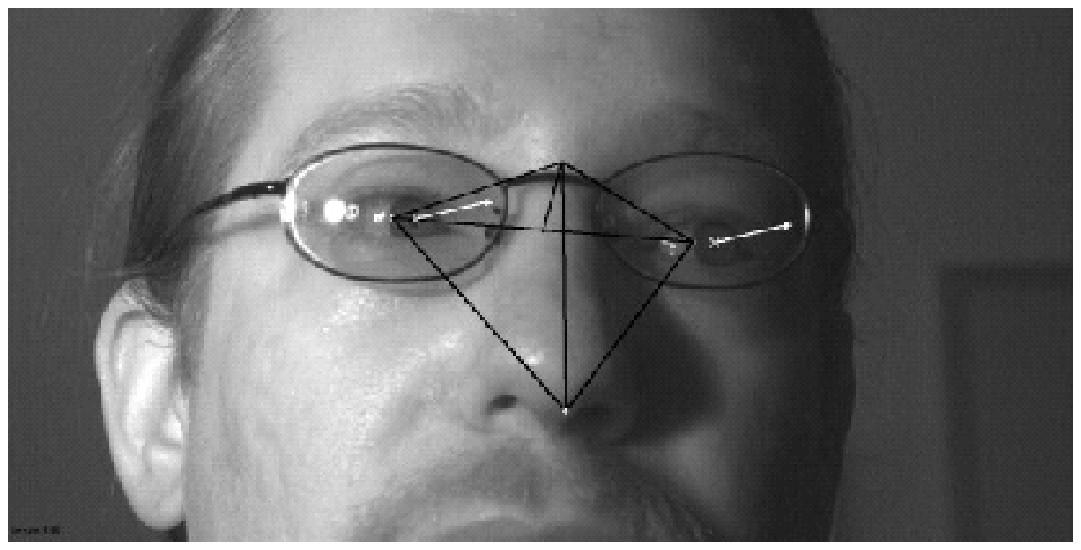,height={8cm}}
\caption{\protect\small The detected triangle, eyes' centers and the nose bottom, together with the gaze line.}
\label{fig::adi3}
\end{figure}

We show sample images produced by the system, where one can see the detected triangle, made of the eyes' centers and the bottom points of the nose. In addition the gaze is reprojected onto the images and rendered by white arrows, figure~\ref{fig::adi1}, \ref{fig::adi2} and \ref{fig::adi3}. 

\section{Discussion}

We proposed an automatic, non-intrusive eye-gaze system. It uses an anthropomorphic model of the human face to calculate the face distance, orientation and gaze angle, without requiring any user-specific calibration. This generality, as seen in subsection~\ref{subsec::robust}, does not introduce large errors into the gaze direction computation. 

While the benefits of a calibration-free system allow for a broad range of previously impossible applications, the system design allows for easy plugging of user-specific calibration data, which will increase the accuracy even more.

\end{document}